\documentclass[a4paper]{article}
\usepackage{multirow}
\usepackage{xcolor}
\usepackage{caption}
\usepackage{subfigure}
\usepackage{url}
\usepackage{tablefootnote}
\usepackage{INTERSPEECH2021}

\title{Language-specific Characteristic Assistance for Code-switching Speech Recognition}
\name{Tongtong Song$^1$, Qiang Xu$^1$, Meng Ge$^{1,2}$, Longbiao Wang$^{1*}$, Hao Shi$^{3*}$, \\Yongjie Lv$^1$, Yuqin Lin$^1$, Jianwu Dang$^{1,4}$\thanks{*Corresponding author.}}
\address{
  $^1$Tianjin Key Laboratory of Cognitive Computing and Application,\\
  College of Intelligence and Computing, Tianjin University, Tianjin, China\\
  $^2$Department of Electrical and Computer Engineering, National University of Singapore, Singapore \\
  $^3$Graduate School of Informatics, Kyoto University, Sakyo-ku, Kyoto, Japan \\
  $^4$Japan Advanced Institute of Science and Technology, Ishikawa, Japan}
\email{\{songtongtong,longbiao\_wang\}@tju.edu.cn, shi@sap.ist.i.kyoto-u.ac.jp}

\begin{document}

\maketitle
\begin{abstract}
Dual-encoder structure successfully utilizes two language-specific encoders (LSEs) for code-switching speech recognition. Because LSEs are initialized by two pre-trained language-specific models (LSMs), the dual-encoder structure can exploit sufficient monolingual data and capture the individual language attributes. However, most existing methods have no language constraints on LSEs and underutilize language-specific knowledge of LSMs. In this paper, we propose a language-specific characteristic assistance (LSCA) method to mitigate the above problems. Specifically, during training, we introduce two language-specific losses as language constraints and generate corresponding language-specific targets for them. During decoding, we take the decoding abilities of LSMs into account by combining the output probabilities of two LSMs and the mixture model to obtain the final predictions. Experiments show that either the training or decoding method of LSCA can improve the model's performance. Furthermore, the best result can obtain up to 15.4\% relative error reduction on the code-switching test set by combining the training and decoding methods of LSCA. Moreover, the system can process code-switching speech recognition tasks well without extra shared parameters or even retraining based on two pre-trained LSMs by using our method.
\end{abstract}

\noindent\textbf{Index Terms}: language-specific characteristic assistance, dual-encoder, code-switching, speech recognition

\section{Introduction}
\begin{figure*}[htbp!]
  \centering
  \includegraphics[width=17cm]{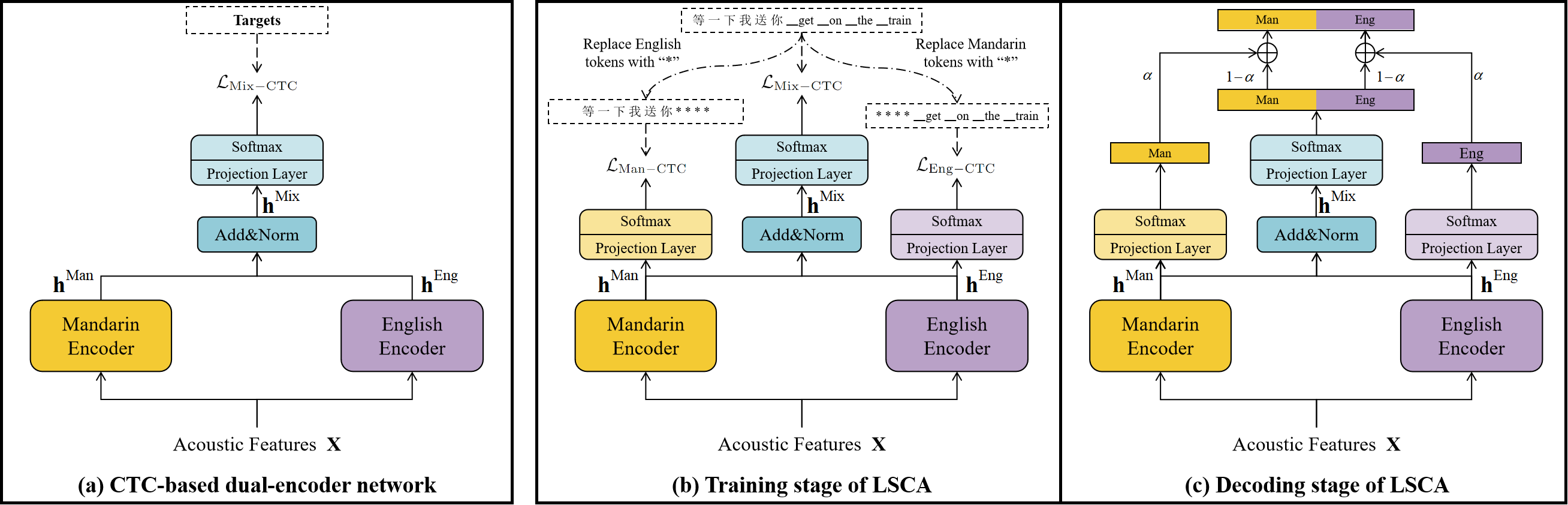}
  \caption{The CTC-based dual-encoder network and the proposed LSCA method. The LSCA method includes the training and decoding stages. In (b), ``*" denotes OOV token ``unk". In (c), we only show the situation of the modeling units belonging to Chinese characters or English BPE.}
  \label{fig:model}
\end{figure*}

Bilingual mixing has become a common phenomenon in international communication. Bilingual mixing, also called code-switching (CS), includes inter-sentential code-switching and intra-sentential code-switching. With the development of automatic speech recognition (ASR) \cite{graves2014towards,graves2013speech,chan2016listen,kim2017joint,watanabe2017hybrid,dong2018speech}, CS ASR is urgently needed and attracts more attention \cite{vu2012first,sitaram2019survey,zeng2018end,shan2019investigating,li2019towards,shah2020learning}.

Compared to monolingual ASR with large monolingual data, CS ASR is limited by hard-to-collect speech and transcriptions. Therefore, reducing the demand for CS data and making full use of monolingual data have become research hotspots \cite{khassanov2019constrained,taneja2019exploiting,lu2020bi,zhang2019towards,zhou2020multi,nj2020investigation}. Dual-encoder structure is an effective way to make full use of monolingual data \cite{lu2020bi,zhang2019towards,zhou2020multi,nj2020investigation}. Under this structure, two language-specific encoders (LSEs) are initialed by corresponding pre-trained language-specific models (LSMs). Then the language-specific features can be extracted and fused by the following shared layers. The high-level representations have strong language discrimination, and it is beneficial for bilingual situation. Therefore, the mixture model achieves comparable performance in two monolingual test sets compared to the well-trained LSMs \cite{lu2020bi,zhang2019towards}. Furthermore, the mixture model can recognize CS sentences without language identity information and CS training data \cite{zhang2019towards}.

However, existing methods based on dual-encoder structure have some problems in the training and decoding stages. In the training stage, only the mixture loss, is used to optimize the entire model. Most dual-encoder methods apply no language constraints on the corresponding LSEs \footnote{We noticed a related work \cite{yan2022joint} was published after submission of our work to Interspeech 2022.}. Then the mixture loss will bias the language-specific features to another language and weaken the strong language discrimination of language-specific features during training. It may decrease the performance of the model in bilingual scenario. And projection layers of two pre-trained LSMs are discarded in existing methods, which does not fully use the language-specific knowledge of two pre-trained LSMs. In the decoding stage, we find that LSMs tend to recognize speech in unseen language as in-vocabulary tokens with low confidence or as out-of-vocabulary (OOV) token ``unk", but have higher confidence when recognizing the speech in the corresponding language. This will be discussed in Section \ref{section:ana}. Therefore, the outputs of LSMs have rich language-specific knowledge. However, only the outputs of the mixture model are used for decoding in existing methods. 

In this work, we propose a language-specific characteristic assistance (LSCA) method to mitigate the above problems in the training and decoding stages. During training, we introduce language-specific losses as language constraints for LSEs and interpolate them with the mixture loss. Moreover, Since the language-specific losses will face the tokens in another language, we generate the language-specific targets for them. During decoding, we take the decoding abilities of two LSMs into account and combine the output probabilities of two LSMs and the mixture model to obtain the final predictions. 

Moreover, we explore the impacts of different weights of the language-specific losses in the whole loss and the output probabilities of two LSMs in the final output probabilities for model performance during training and decoding, respectively. In addition, we also explore completing CS ASR tasks by adding no extra shared layers or even without retraining based on two pre-trained LSMs.

The rest of this paper is organized as follows. Section 2 describes the CTC-based dual-encoder network for CS ASR as our baseline system. In Section 3, we propose the LSCA method improves the CTC-based dual-encoder network in the training and decoding stages. The experiment details are presented, and the proposed method is evaluated in Section 4. Finally, we conclude this paper and discuss the future work.

\section{CTC-based dual-encoder network for CS ASR}\label{section:baseline}

Transformer \cite{vaswani2017attention} has been introduced into ASR and received comparable performance to the conventional hybrid and other end-to-end approaches \cite{dong2018speech,nakatani2019improving}. It is also widely used in CS ASR tasks \cite{lu2020bi,zhang2019towards,zhou2020multi,nj2020investigation,zhang2021end,zhang2021decoupling,huang2021improving}. Because the encoder of the Transformer already has strong language modeling capability, it can achieve comparable performance by only using Connectionist Temporal Classification (CTC)  \cite{graves2006connectionist} without the decoder during decoding \cite{zhang2020unified,yao2021wenet}. Since the dual-encoder structure can better use large amounts of monolingual data and capture the individual language attributes, it can perform better than the single-encoder structure in bilingual scenario \cite{lu2020bi,zhang2019towards,zhou2020multi,nj2020investigation}.

In this work, we use a CTC-based dual-encoder network to leverage large monolingual data. The encoder of the Transformer is used as the feature extractor, and CNN is used to do time-scale down-sampling for the input acoustic features before each encoder, and CTC \cite{graves2006connectionist} loss is the objective function. 

As shown in Figure \ref{fig:model} (a), two LSEs named Mandarin Encoder and English Encoder are initialized by two pre-trained LSMs, respectively. This two pre-trained LSMs are trained with large scale monolingual data, they have learned language-specific knowledge. Therefore, language-specific features $\mathbf{h}^{\rm Man}$ and $\mathbf{h}^{\rm Eng}$ can be extracted from the input acoustic features $\mathbf{x}$, and they have strong language discrimination. Then, they are added and through a layer-normalized affine transformation to get the mixture features $\mathbf{h}^{\rm Mix}$:
\begin{gather}
     \mathbf{h}^{\rm Man}=MandarinEncoder(\mathbf{x})\\
     \mathbf{h}^{\rm Eng}=EnglishEncoder(\mathbf{x}) \\
     \mathbf{h}^{\rm Mix}=LayerNorm(\mathbf{h}^{\rm Man}+\mathbf{h}^{\rm Eng})
\end{gather}

The mixture features ${\mathbf{h}^{\rm Mix}}$ are input to the mixture projection and softmax layer, and then calculate the mixture loss ${\mathcal{L}_{\rm Mix-CTC}}$ with original targets. The entire model is finally optimized using a small amount of CS data with ${\mathcal{L}_{\rm Mix-CTC}}$.

For modeling units, we use Chinese characters for Mandarin and BPE \cite{sennrich2015neural} for English represented by $\phi_{\rm character}$ and $\phi_{\rm BPE}$, respectively. Furthermore, they are combined to generate the Character-BPE modeling units for the mixture model.

\section{Language-specific characteristic assistance}
This section describes the proposed LSCA method that improves the CTC-based dual-encoder network. The proposed method is shown in the right part of Figure \ref{fig:model}. It includes the training stage in Figure \ref{fig:model} (b) and the decoding stage in Figure \ref{fig:model} (c). They can be used alone or in combination.
\subsection{The training stage of the LSCA method}
In the training stage, in order to add language constraints to the LSEs, we take two language-specific CTC losses (LSCs) into account and interpolate them with the mixture loss ${\mathcal{L}_{\rm Mix-CTC}}$ as the new objective function \footnote{We noticed a related work \cite{yan2022joint} was published right after submission of our work to Interspeech 2022. However, they use ``Man" and ``Eng" to mask Chinese and English tokens, respectively, instead, we use ``unk" for both.}: 
\begin{gather}
  \mathcal{L}=(1-\lambda)\mathcal{L}_{\rm Mix-CTC}+\lambda\mathcal{L}_{\rm LS-CTC} \\
  \mathcal{L}_{\rm LS-CTC}=\frac{\mathcal{L}_{\rm Man-CTC}+\mathcal{L}_{\rm Eng-CTC}}{2}
\end{gather}
\noindent{where $\mathcal{L}_{\rm LS-CTC}$ is the combination of two LSCs $\mathcal{L}_{\rm Man-CTC}$ and $\mathcal{L}_{\rm Eng-CTC}$, represent the CTC loss for Mandarin encoder and English encoder, respectively. $\lambda\in[0,1]$ is the weight of $\mathcal{L}_{\rm LS-CTC}$ in the whole loss. Specially, $\lambda=0$ and $\lambda=1$ means only training with ${\mathcal{L}_{\rm Mix-CTC}}$ and $\mathcal{L}_{\rm LS-CTC}$, respectively. When $\lambda=0$, the system is the same as the baseline. When $\lambda=1$, there are no extra parameters based on two pre-trained LSMs.}

Since we reuse the LSCs of two pre-trained LSMs, the corresponding projection layers have learned language-specific knowledge. However, in the training stage, each LSC will face some sentences with the tokens in another language, and these tokens need to be shielded to prevent LSMs from learning knowledge of another language. For this purpose, we need to generate language-specific targets for LSCs. As shown in Figure \ref{fig:model} (b), the language-specific targets are generated by mapping the tokens in another language to the OOV token ``unk" based on the original targets. Then the LSMs will focus on the parts in the corresponding language and ignore the other parts of the speech just like human beings. ${\mathcal{L}_{\rm Mix-CTC}}$ uses the mixture vocabulary while LSCs use the corresponding vocabulary.

\subsection{The decoding stage of the LSCA method}
In the decoding stage, the LSMs have rich language-specific knowledge, especially after the training stage of our method. In order to make full use of the language-specific knowledge for better decoding, we take the decoding abilities of LSMs into account by combining the output probabilities of LSMs and the mixture model to obtain the final predictions.

However, the output dimensions of two LSMs and the mixture model are not the same. We can not directly add the output probabilities of these three parts. In addition, we found that the probabilities of several modeling units with relatively high probability account for a large proportion. Therefore, it is reasonable to add the corresponding parts of the modeling units proportionally.

When the modeling unit belongs to $\phi_{\rm character}$ or $\phi_{\rm BPE}$, the final probability will be obtained by interpolating the outputs of two parts, the corresponding LSMs and the corresponding part of the mixture model like Figure \ref{fig:model} (c). Since two LSMs and the mixture model can accurately predict the token ``blank", the final probability of ``blank" is obtained by interpolating the outputs of these three parts. Because we use ``unk" as unseen language token in LSMs, so the token ``unk" for the mixture model is true OOV token, and it has little impact on the system performance:
\begin{equation}
\label{eq:decoing}
\begin{normalsize}
P_u=
 \begin{cases}
    (1-\alpha)P^{\rm Mix}_u+\alpha P^{\rm Man}_u \hfill u\in\phi_{\rm character} \\
    (1-\alpha)P^{\rm Mix}_u+\alpha P^{\rm Eng}_u \hfill u\in\phi_{\rm BPE}\\
    (1-\alpha)P^{\rm Mix}_u+\alpha\frac{P^{\rm Eng}_u+P^{\rm Man}_u}{2} \hfill \quad u=blank
 \end{cases}
\end{normalsize}
\end{equation}
\noindent{where $u$ represents a modeling unit, $P^{\rm Mix}_u$, $P^{\rm Man}_u$ and $P^{\rm Eng}_u$ represent the probabilities of $u$ in the output probabilities of the mixture model and two LSMs, respectively. $\alpha\in[0,1]$ controls the weight of the output probabilities of LSMs in the final output probabilities. Moreover, $\alpha=0$ means that only the mixture model are used for decoding. $\alpha=1$ means that only the LSMs are used for decoding.}
\section{Experiments}
\subsection{Data}
Our experiments have two monolingual corpora and one Mandarin-English CS corpus:\\   
\textbf{Monolingual corpora:} (1) AISHELL-2 \cite{du2018aishell} with 1K hours monolingual Mandarin read speech. (2) Librispeech \cite{panayotov2015librispeech} with 960 hours monolingual English read speech.
\\
\textbf{CS corpus:} The Mandarin-English CS corpus, which is from a contest \footnote{http://contest.aicubes.cn}, contains not only Mandarin-English code-switching but also monolingual sentences. It covers daily conversation, tourism, finance, and other fields. The sentences are recorded with a microphone in the silent room or a mobile phone in the office, and in single-channel, 16 kHz, 16-bit PCM format. The information of its training and test sets are described in Table \ref{tab:cs_data}.

\begin{table}[th]
\setlength\tabcolsep{1.5mm}{
\caption{The details of training and test sets of the Mandarin-English CS corpus. M-E means Mandarin-English code-switching.}
\label{tab:cs_data}
\centering
  \begin{tabular}{llcccc}
    \hline
    \hline
     Set  & Category & Mandarin & English & M-E & SUM\\
    \hline
    \multirow{3}{*}{Train}    & Ratio  & 26.75\%  & 25.08\%  & 48.17\%  & 100\%\\
                              & \# Utterance  & 2140  & 2006  & 3854  & 8000\\
                              &  Duration (h)  & 1.92  & 3.50  & 5.32  & 10.74\\
    \hline
    \multirow{3}{*}{Test}     & Ratio  & 20.09\%  & 32.41\%  & 47.50\%    & 100\%\\
                              & \# Utterance  & 3416  & 5509  & 8075  & 17000\\
                              & Duration (h)  & 3.05  & 9.63  & 11.26  & 23.94\\
  \hline
  \hline
  \end{tabular}}
\end{table}

\subsection{Experimental setup}
We extract 80-dimensional log-Mel filterbanks with a window size of 25 ms and a step size of 10 ms as the acoustic features. SpecAugment \cite{park2019specaugment} is applied with 2 frequency masks (F = 10) and 3 time masks (T = 50) during all training stages.

Each LSE has a 12-layer Transformer encoder with attention dimension $d_{model}=256$ and the feedforward network dimension $d_{ffn}=1024$. And 4 heads are used for multi-head attention. 2-layer CNN is used to down-sample the time dimension of the input features to one quarter. All models are trained for 50 epochs, and warmup \cite{vaswani2017attention} is used for ﬁrst 250K iterations for pre-training two monolingual models, and 2500 for training with CS data, because of the smaller data scale of CS training data. The dropout is 0.1 to avoid overfitting. The maximum number of frames in one batch is 10K. The last 5 checkpoints are averaged and greedy search is used for decoding.

There are about 3K Chinese characters and 5K English BPE for the modeling units, OOV character or BPE units are mapped to ``unk" token. We report a mix error rate (MER) for the CS test set with character error rate (CER) for the Mandarin part and word error rate (WER) for the English part.
\begin{table}[th] 
\caption{Performance (MER\%) of different systems on the CS test set. The CTC-based dual-encoder network is our baseline system.}
\label{tab:results1}
\centering
\setlength\tabcolsep{3.0mm}{
  \begin{tabular}{lcc}
     \hline
     \hline
   Type & System & MER \\
    \hline
  Hybrid & Kaldi(sMBR)  & 38.71 \\
   \hline
  \multirow{2}{*}{E2E} &  CTC-based single-encoder network & 29.18 \\
   & CTC-based dual-encoder network & \textbf{27.35} \\
     \hline
     \hline
  \end{tabular}}
\end{table}
\subsection{Evaluation of baseline system}
There are three different systems, as shown in Table \ref{tab:results1}, including one Hybrid and two E2E systems. The Kaldi system \cite{povey2011kaldi} is from the official organizers of the contest \footnote{https://github.com/10jqka-aicubes/code-switching-contest}. The Kaldi system first is trained with about 150 hours Mandarin read speech AISHELL-1 \cite{bu2017aishell} and 100 hours English read speech $train\_clean\_100$ of Librispeech \cite{panayotov2015librispeech} with 3-ways speed perturbation, then the model is trained with the CS training data and sMBR criterion. In addition, We also train a CTC-based single-encoder network, and the other sets are the same as the CTC-based dual-encoder network. The MER of the CTC-based dual-encoder network is 27.35\%, which performs better than the other two systems and shows it is a strong baseline system for the experiments.

\subsection{Comparative study on the LSCA method}

In Table \ref{tab:results2}, each column shows the impacts of different weights $\lambda$ of LSCs on model performance in the training stage of LSCA. Each row shows the impacts of different weights of LSMs output probabilities on decoding results in the decoding stage of LSCA. When $\lambda=0$ and $\alpha=0$, it is the same as the baseline system.
\begin{table}[th]
\caption{The impacts of different values of $\lambda$ and $\alpha$ on the model performance (MER\%). “-” means the results are unavailable for those sets.}
\label{tab:results2}
\centering
  \setlength\tabcolsep{1.7mm}{
  \begin{tabular}{cccccccc}
  \hline
  \hline
  \multirow{2}{*}{$\lambda$} & \multicolumn{7}{c}{$\alpha$} \\
  \cline{2-8}
  & 0  & 0.1  & 0.3  & 0.5  & 0.7  & 0.9  & 1.0 \\
  \hline
  0                 & 27.35  & 27.02  & 25.98  & 25.56  & 30.30  & 34.69  & 36.50  \\
  \hline
  0.1               & 26.56  & 26.25  & 25.34  & 24.31  & 24.05  & 24.32  & 24.54  \\
  0.3               & 26.14  & 25.81  & 24.91  & 23.91  & 23.49  & 23.60  & 23.68  \\
  0.5               & 26.01  & 25.67  & 24.74  & 23.75  & 23.29  & 23.31  & 23.39  \\
  0.7               & 25.87  & 25.52  & 24.63  & 23.69  & \textbf{23.13}  & 23.16  & 23.27  \\
  0.9               & 26.34  & 25.92  & 24.86  & 23.77  & 23.23  & 23.24  & 23.31  \\
  \hline
  1.0               & -      & -      & -      & -      &  -     & -      & 23.57  \\
  \hline
  \hline
  \end{tabular}}
\end{table}

When $\alpha=0$, only the training stage changes to the LSCA method, the decoding stage is the same as the baseline system. The system can achieve up to 5.4\% relative error reduction over the baseline system at $\lambda=0.7$. This proves that adding language constraints to LSEs can improve the acoustic representation of each language and improves the outputs of the mixture model. 

When $\lambda=0$, the training stage is the same as the baseline system, only the decoding stage changes to the LSCA method. The system can obtain up to a relative 6.5\% error reduction at $\alpha=0.5$. This shows that we better take advantage of language-specific knowledge of LSMs to improve the decoding results.

Furthermore, the system greatly improves when the training and decoding stages change to the LSCA method. We can get the best result at $\lambda=0.7$ and $\alpha=0.7$, and the relative error reduction is up to 15.4\% compared to the baseline system. This proves that our decoding method can better combined with our training method, and fully demonstrates the effectiveness of our method.

When $\alpha=1$, only the outputs of LSMs are used for decoding, the system can achieve good performance without using the mixture model for decoding. Therefore, we argue that even without any extra parameters based on two pre-trained LSMs, the system still performs well by using the proposed LSCA method for CS ASR. Then, we let $\lambda=1$ and $\alpha=1$, only training the model with LSCs, and decoding with the outputs of LSMs. The relative error reduces up to 13.8\% compared to the baseline system. Therefore, we think the system can also perform well for CS ASR, even directly decoding on two pre-trained monolingual models by using our decoding method, and we get 47.84\% MER on the CS test, which provides a good solution for directly using large-scale pre-trained monolingual models for CS ASR.

\subsection{Analyse the outputs of two LSMs}\label{section:ana}
\begin{figure}[htbp]
  \centering
  \includegraphics[width=8cm]{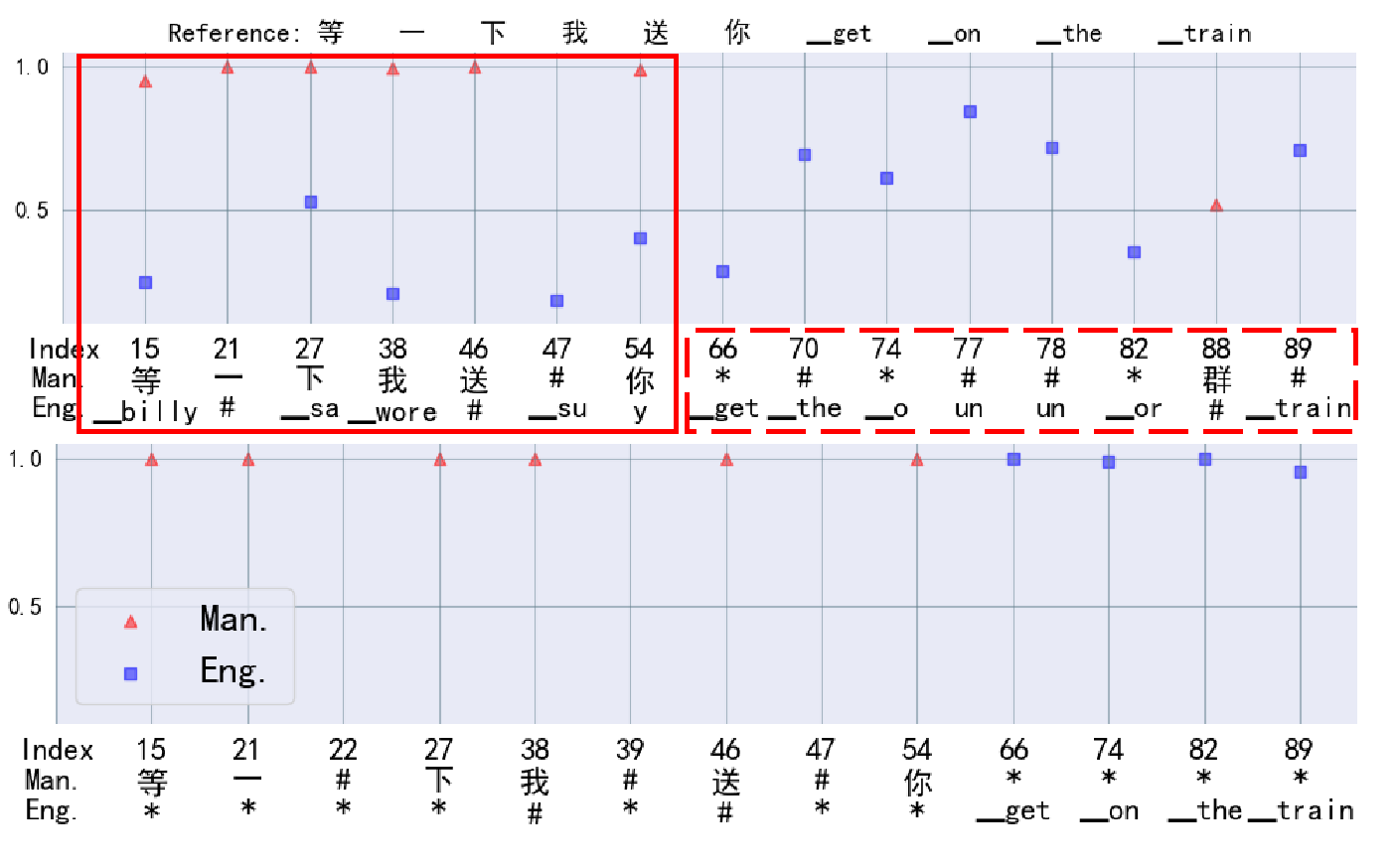}
  \caption{Visualization of the top one probabilities and the decoding results of two LSMs. The upper part belongs to two pre-trained monolingual models. The under part belongs to two LSMs after the training stage of the LSCA method with $\lambda=0.7$. For better display, we denote ``unk" and ``blank" tokens as ``$*$" and ``$\#$" respectively. We remove those frames which are both ``blank" predicted by two models. Furthermore, we do not display the probabilities of ``unk". ``Index" means frame index in the outputs of one utterance. ``Man." and ``Eng." mean the predictions of Mandarin and English models, respectively.}
  \label{fig:sample2}
\end{figure}

We visualize the top one probabilities and the decoding results of two LSMs before and after the training stage of LSCA. As shown in the upper part of Figure \ref{fig:sample2}, the monolingual model has higher confidence when the speech is in the corresponding language, like the triangle markers in the solid box, but has lower confidence when in unseen languages, like the square markers in the solid box. Moreover, monolingual models tend to map the speech in unseen languages to the OOV token, like the Mandarin predictions in the dotted box.

As shown in the under part of Figure \ref{fig:sample2}, after the training stage of the LSCA method, LSMs have higher confidence when the speech is in the corresponding language, like the triangle and square markers. Because we map the tokens in another language to ``unk" based on the original targets in the training stage of the LSCA method. Then, LSMs predict the tokens in another language to ``unk" with high confidence, like the predictions of Mandarin model and English model. This fully demonstrates the effectiveness of the proposed LSCA method.

\section{Conclusion and future work}

In this paper, we proposed a LSCA method that improves the training and decoding stages based on the CTC-based dual-encoder network for CS ASR. The proposed training and decoding method can work alone or in combination. Experimental results show that we can obtain up to 15.4\% relative error reduction compared to the baseline system by combining the proposed training and decoding method. In addition, the system processes CS ASR tasks well without adding extra parameters or even retraining based on two monolingual pre-trained models by using the proposed method. This provides a good solution for quickly and directly using large-scale monolingual models for CS ASR.

For future work, we will further try our method on larger CS corpus. For model structure, we will introduce our method into the encoder-decoder structure. We will try to apply our method to multilingual speech recognition.

\section{Acknowledgements}
This work was supported in part by the National Natural Science Foundation of China under Grant 62176182 and Alibaba Group through Alibaba Innovative Research Program.
\clearpage
\bibliographystyle{IEEEtran}

\bibliography{mybib}


\end{document}